\documentclass[conference]{IEEEtran}
\IEEEoverridecommandlockouts
\usepackage{cite}
\usepackage{amsmath,amssymb,amsfonts}
\usepackage{algorithmic}
\usepackage{graphicx}
\usepackage{textcomp}
\usepackage{xcolor}
\usepackage{booktabs}
\usepackage{multirow} 
\usepackage{enumitem}
\usepackage{array}
\usepackage[caption=false,font=footnotesize]{subfig}
\def\BibTeX{{\rm B\kern-.05em{\sc i\kern-.025em b}\kern-.08em
    T\kern-.1667em\lower.7ex\hbox{E}\kern-.125emX}}
\begin{document}

\title{Explicit Knowledge-Guided In-Context Learning for Early Detection of Alzheimer’s Disease\\
}

\author{
    Puzhen Su,
    Yongzhu Miao,
    Chunxi Guo,
    Jintao Tang\textsuperscript{*},
    Shasha Li\textsuperscript{*},
    Ting Wang\textsuperscript{*}\thanks{*Corresponding author.} \\
    College of Computer Science and Technology, National University of Defense Technology, Changsha, China \\
    \texttt{\{supuzhen, miaoyz, chunxi, tangjintao, shashali, tingwang\}@nudt.edu.cn}
}

\maketitle

\begin{abstract}
Detecting Alzheimer’s Disease (AD) from narrative transcripts remains a challenging task for large language models (LLMs), particularly under out-of-distribution (OOD) and data-scarce conditions. While in-context learning (ICL) provides a parameter-efficient alternative to fine-tuning, existing ICL approaches often suffer from task recognition failure, suboptimal demonstration selection, and misalignment between label words and task objectives, issues that are amplified in clinical domains like AD detection. We propose Explicit Knowledge In-Context Learners (EK-ICL), a novel framework that integrates structured explicit knowledge to enhance reasoning stability and task alignment in ICL. EK-ICL incorporates three knowledge components: confidence scores derived from small language models (SLMs) to ground predictions in task-relevant patterns, parsing feature scores to capture structural differences and improve demo selection, and label word replacement to resolve semantic misalignment with LLM priors. In addition, EK-ICL employs a parsing-based retrieval strategy and ensemble prediction to mitigate the effects of semantic homogeneity in AD transcripts. Extensive experiments across three AD datasets demonstrate that EK-ICL significantly outperforms state-of-the-art fine-tuning and ICL baselines. Further analysis reveals that ICL performance in AD detection is highly sensitive to the alignment of label semantics and task-specific context, underscoring the importance of explicit knowledge in clinical reasoning under low-resource conditions.
% This document is a model and instructions for \LaTeX.
% This and the IEEEtran.cls file define the components of your paper [title, text, heads, etc.]. *CRITICAL: Do Not Use Symbols, Special Characters, Footnotes, 
% or Math in Paper Title or Abstract.
\end{abstract}

\begin{IEEEkeywords}
Alzheimer's Disease, Contextual Learning, LLMs, Explicit Knowledge.
\end{IEEEkeywords}

\section{Introduction}

Alzheimer’s disease (AD) is a global health crisis, causing profound cognitive decline and significant socio-economic impact\cite{AD_bg_1}. Early detection is critical for effective intervention, yet traditional diagnostic methods like cerebrospinal fluid (CSF) analysis\cite{CSF} and neuroimaging\cite{Neuroimaging} are invasive and costly. In response, non-invasive, text-based approaches\cite{roshanzamir2021da}, that analyze linguistic patterns have emerged as a promising alternative.% zhu2021dlm, ilias2023context

\begin{figure}[!ht]
\centering
\small
\includegraphics[width=0.85\linewidth]{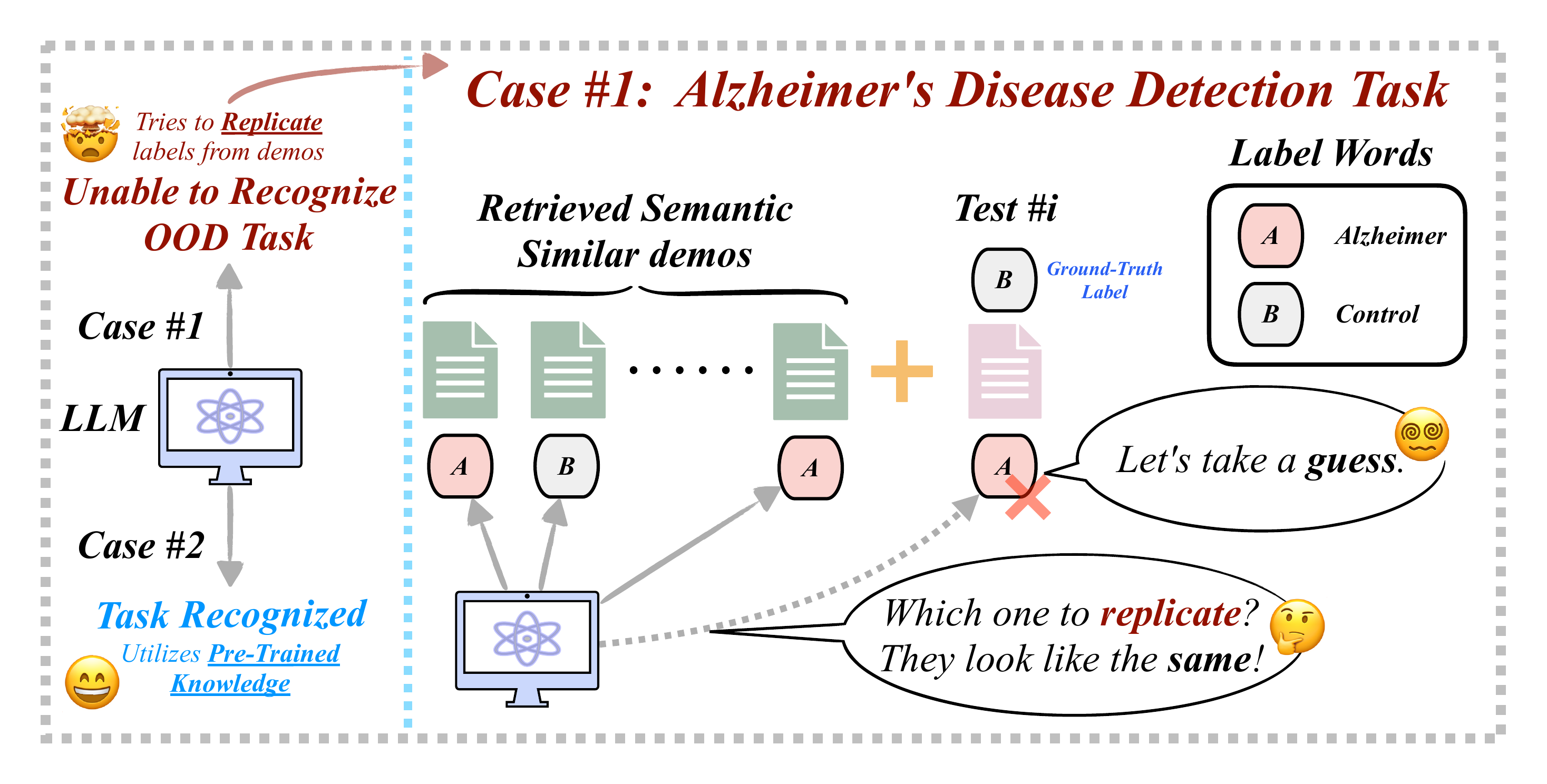}
\vspace{-0.4cm}
\caption{
\textbf{Schematic illustration of reasoning failure in ICL under out-of-distribution (OOD) task scenarios.}
}\label{figure:0}
\vspace{-0.5cm}
\end{figure}

Current AD detection methods predominantly rely on adapting small language models (SLMs) for specific tasks. While effective, SLMs methods are limited by their dependency on extensive task-specific data, leading to over-specialization and pre-trained knowledge forgetting\cite{forget_1}, 
%forget_2,forget_3
which hampers generalization to unseen data, especially in out-of-distribution (OOD) scenarios. While data augmentation (DA) techniques\cite{spz} improve performance, they still rely heavily on data alignment and introduce bias due to random operation. In contrast, large language models (LLMs) such as LLaMA and GPT-4, with their remarkable in-context learning (ICL) capabilities, offer a promising solution. ICL %\cite{prompt_learning_1}  prompt_learning_2 \cite{ICL_example_1}
enables LLMs to adapt to new tasks with minimal data, using in-context demos without any parameter updates. However, existing ICL methods have underperformed in AD detection, with even advanced models like GPT-4\cite{ICL_In_AD} failing to surpass traditional SLMs approaches. 
As illustrated in Fig.~\ref{figure:0}, when LLMs encounter tasks they have not been pre-trained for, the absence of semantic contrast among retrieved demos combined with semantically unrelated label words leads to unreliable reasoning, prompting LLMs to replicate demo labels rather than engage in context-informed decision making.

\begin{figure*}[!htbp]
\centering
\small
\includegraphics[width=0.85\linewidth]{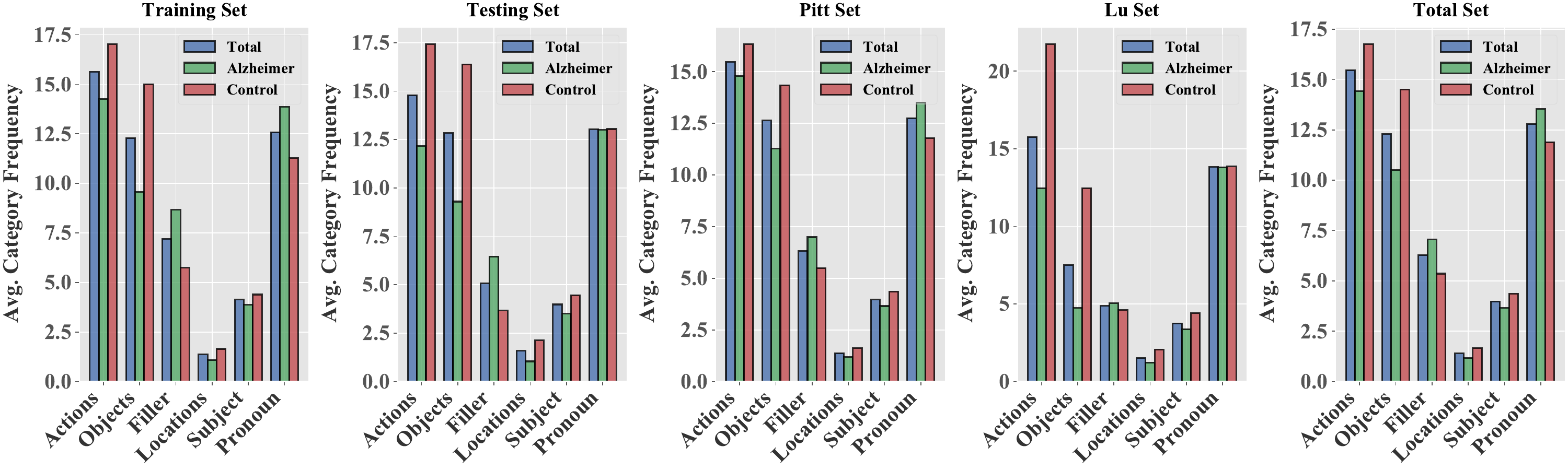}
\vspace{-0.4cm}
\caption{
\textbf{Parsing category distributions across datasets in AD detection.}
This figure compares the average frequency of six syntactic categories—\textit{Actions, Objects, Fillers, Locations, Subjects,} and \textit{Pronouns}—across the ADReSS challenge-Train, ADReSS challenge-Test, Lu, Pitt, and Total datasets.
}\label{figure:distribution}
\vspace{-0.2cm}
\end{figure*}

Such shortfall can be attributed to two interrelated factors, \textbf{demos quality} and \textbf{task recognition}\cite{ICL_unveiling}, which are keys to effective ICL. First, the quality of retrieved demos is limited in AD detection due to transcript-specific characteristics. Most ICL methods construct input contexts by retrieving demos either randomly or based on semantic similarity. However, these strategies struggle due to the \textbf{\textit{semantic homogeneity}} of transcripts, where both AD and control participants describe the same picture. Additionally, as shown in Fig.~\ref{figure:distribution}, different categories of transcripts exhibit distinct parsing structures, yet, conventional ICL methods fail to leverage such \textbf{\textit{parsing variation}}, limiting their ability to retrieve informative demos. Second, effective task recognition depends on how well the task objective is semantically aligned with both transcript and label words. In AD detection, the transcripts are descriptive narratives rather than direct indicators of cognitive impairment, making it difficult to establish a strong semantic connection with the task objective. Furthermore, the semantic misalignment between label words and task can lead to severe performance drops in ICL\cite{Label_words_1}. Thus, when conventional label words fail to align with the nuanced linguistic features of AD transcripts, the difficulty of task recognition exacerbates.%(\textit{Alzheimer}/\textit{Control}) %demo_sem_2 || Label_words_2  \cite{Random_demo} \cite{demo_sem_1}

To address the challenges of existing ICL methods, we propose EK-ICL (\textbf{E}xplicit \textbf{K}nowledge \textbf{I}n-\textbf{C}ontext \textbf{L}earners), a novel framework designed to enhance LLMs reasoning in AD detection tasks. EK-ICL integrates three forms of explicit knowledge: 1) \textbf{Confidence scores}, derived from SLM logits, which provide reliable insights into prediction and stabilize task recognition; 2) \textbf{Parsing feature scores}, which highlight linguistic differences between samples and serve as the basis of improving demo retrieval quality that captures structural distinctions; and 3) \textbf{In-distribution (ID) label words}, which have been investigated for this task, and replacing traditional task-specific label words with ID labels to better align with the pre-trained knowledge of LLMs. Additionally, EK-ICL employs an ensemble-based ICL strategy that aggregates predictions through majority voting to mitigate label conflict. We demonstrate that EK-ICL effectively addresses the limitations of existing ICL methods and outperforms SOTA SLMs and LLMs methods across three AD detection datasets. Our contributions can be summarized as follows:
\begin{itemize}[nosep]
\item We propose EK-ICL, a novel ICL framework for AD detection that integrates explicit knowledge to enhance task recognition and demo selection.
\item EK-ICL captures AD-specific linguistic patterns via SLM-derived confidence and parsing-based feature scores, improving reasoning under OOD and low-resource settings.
\item We conduct a systematic study on label word configurations, revealing that task-label alignment is critical for robust ICL in clinical AD detection.
\end{itemize}

\begin{figure}[!t]
\centering
\small
\includegraphics[width=0.95\linewidth]{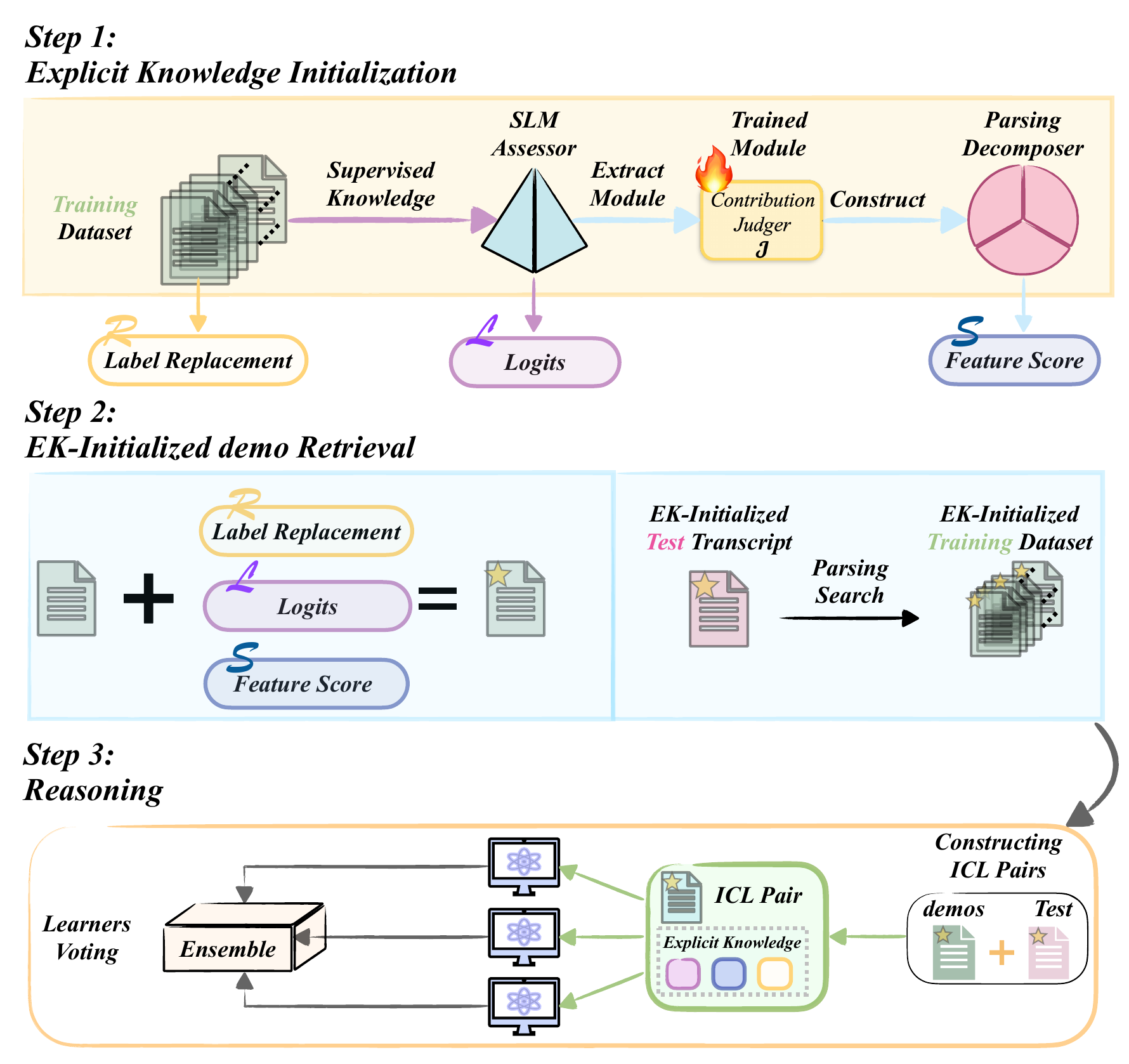}
\caption{
\textbf{Overview of the proposed EK-ICL framework.}
}\label{figure:framework}
\end{figure}

% This document is a model and instructions for \LaTeX.
% Please observe the conference page limits. For more information about how to become an IEEE Conference author or how to write your paper, please visit   IEEE Conference Author Center website: https://conferences.ieeeauthorcenter.ieee.org/.

% \subsection{Maintaining the Integrity of the Specifications}

% The IEEEtran class file is used to format your paper and style the text. All margins, 
% column widths, line spaces, and text fonts are prescribed; please do not 
% alter them. You may note peculiarities. For example, the head margin
% measures proportionately more than is customary. This measurement 
% and others are deliberate, using specifications that anticipate your paper 
% as one part of the entire proceedings, and not as an independent document. 
% Please do not revise any of the current designations.

\section{Methods}
We propose EK-ICL, a framework that enhances AD detection by integrating explicit knowledge into ICL. As shown in Fig.~\ref{figure:framework}, EK-ICL improves task recognition, demo quality, and label-task alignment by incorporating three forms of explicit knowledge obtained from \textbf{explicit knowledge initialization}. First, \textit{confidence scores} ($\textbf{S}_{\text{conf}}$) are derived from 
SLMs logits and align the transcript with the task, addressing task recognition issues. Second, \textit{parsing feature scores} ($\textbf{S}_{\text{feat}}$) capture syntactic differences between samples, which improves demo quality and task alignment by prioritizing structural similarity. Finally, \textit{ID label replacement} substitutes traditional task-specific labels (i.e., \textit{Alzheimer/Control}), with in-distribution labels (i.e., \textit{Good/Bad}), ensuring better alignment with the task. These components are further complemented by \textbf{parsing search}, which enhances demo selection by leveraging parsing similarity, and \textbf{ensemble in-context learners}, which aggregates predictions from multiple learners using majority voting, addressing label conflict and stabilizing predictions. The following sections describe each component in detail.

\subsection{Explicit Knowledge Initialization}
\paragraph{SLM Assessor (EK-I Contribution Scores)}  
To quantify model confidence and stabilize reasoning, we incorporate a lightweight SLM with two modules: a noise \textit{\textbf{Injector}} that perturbs token representations and a contribution \textit{\textbf{Judger}} that evaluates token-level contributions.

Given a token representation $\mathbf{e}_i$ in transcript $T_j$, the \textit{\textbf{Injector}} produces a noise representation $\mathbf{e}'_i$ using the Gumbel-Softmax trick \cite{jang2016categorical}. 

The \textit{\textbf{Judger}} then assigns a token-specific contribution weight $p_i$ via a linear probe with sigmoid activation:
\begin{equation}
\small
p_i=\sigma\!\left(\mathbf{w}^\top \mathbf{e}_i + b\right), \quad p_i\in(0,1).
\end{equation}

Finally, the perturbed token is a convex combination of the original and noise representations:
\begin{equation}
\small
\Tilde{\mathbf{e}}_i = p_i\,\mathbf{e}_i + (1-p_i)\,\mathbf{e}'_i .
\end{equation}

The final text representation, $\Tilde{\mathbf{E}} = \{\Tilde{\mathbf{e}}_1, \Tilde{\mathbf{e}}_2, \cdots, \Tilde{\mathbf{e}}_n\}$, is aggregated using a global max-pooling operation. The logits of the SLM, used as confidence scores $\textbf{S}_{\text{conf}}$, are computed as:
\begin{align}
\small
\mathbf{z} &= \textrm{GlobalMaxPooling}(\Tilde{\mathbf{E}}), \\
\textbf{S}_{\text{conf}} &= \textrm{Sigmoid}(\textrm{MLP}(\mathbf{z})),
\end{align}
where \textrm{MLP} is the multi-layer perceptron. The trained \textbf{\textit{Judger}} outputs are also used to compute feature scores $\textbf{S}_{\text{feat}}$ and parsing similarities.

\paragraph{Parsing Decomposer (EK-II Feature Scores)}  
The parsing decomposer,
%illustrated in Figure~\ref{figure:Decomposer}, 
extracts syntactic structure-based knowledge from AD transcripts, providing explicit parsing feature scores $\textbf{S}_{\text{feat}}$. Differs from previous retrieval approaches that rely on semantic similarity, our method quantifies the contribution of different parsing categories to improve structural differentiation between AD and HC groups. To achieve this, we define six key parsing categories, including \textit{Subject}, \textit{Object}, \textit{Action}, \textit{Location}, \textit{Filler}, and \textit{Pronoun}.

Given a transcript $T_j$ with token-level representations $\mathbf{E_j} = \{\mathbf{e}_{1}, \mathbf{e}_{2}, \cdots, \mathbf{e}_{n}\}$, we first compute the parsing contribution weight $\omega_{j}^k$ for each category $Cat_k$:
\begin{equation}
\small
\omega_{j}^k = \sum_{i=1}^{n} p_{i} \cdot \mathbb{I}(\text{Cat}(\mathbf{e}_{i}) = Cat_k),
\end{equation}
where $\mathbb{I}(\cdot)$ is the indicator function, determining if token $\mathbf{e}_{i}$ belongs to category $Cat_k$, and $p_{i}$ is the output of the \textit{Judger} module, indicating the token's contribution.

Next, we define the contribution array $R(T_j)$ for each transcript $T_j$:
\begin{equation}
\small
R_j = \{\omega_{j}^a, \cdots, \omega_{j}^b|\omega_{j}^a > \omega_{j}^b\}.
\end{equation}

To establish a global ranking, we compute the standard contribution array $R_{\text{std}}$ across the training dataset $\mathcal{D}$:
\begin{equation}
\begin{aligned}
\small
R_{\text{std}} &= \{\Bar{\omega}^a, \cdots, \Bar{\omega}^b \mid \Bar{\omega}^a > \Bar{\omega}^b\},\\
\bar{\omega}^k &= \frac{1}{|\mathcal{D}|} \sum_{x \in \mathcal{D}} \sum_{i=1}^{n_x} p_i \cdot \mathbb{I}\left(\text{Cat}(\mathbf{e}_i) = \text{Cat}_k\right),  
\end{aligned}
\label{eq:avg-weight}
\end{equation}
and then, we can derive the standard category rank:
\begin{equation}
\small
C_{\text{std}}= \mathrm{\arg sort}(R_{std}).
\end{equation}

Finally, the feature scores $\textbf{S}_{\text{feat}}$ for transcript $T_j$ is computed as a weighted sum:
\begin{equation}
\small
\textbf{S}_{\text{feat}} = R_{std}(T_j) \cdot \text{Softmax}(R_{\text{std}}),
\end{equation}
here, $R_{std}(T_j)$ represents the category-wise contribution array of $T_j$, ordered according to $C_{\text{std}}$.

By computing these parsing-based feature scores $\textbf{S}_{\text{feat}}$, we provide explicit structural knowledge that are later used in parsing search to improve demo selection. These scores help distinguish between AD and HC transcripts by emphasizing \textbf{\textit{parsing variations}} rather than relying on semantic similarity.

\paragraph{In-Distribution Labels Replacement (EK-III ID Label Words)}  
Noticing label sensitivity in OOD tasks, we replace default labels (i.e., \textit{Alzheimer}/\textit{Control}) with ID labels (i.e., \textit{Good}/\textit{Bad}), aligning with the LLMs' pre-trained knowledge. This alignment enables LLMs to better leverage their pre-trained understanding of universal concepts, mitigating performance degradation caused by mismatched label semantics in OOD scenarios.

\subsection{Parsing Search and Ensemble Prediction}

\paragraph{Parsing Search}  
Unlike current ICL methods that retrieve demos based on semantic similarity, we introduce parsing similarity, which better captures structural differences in transcripts. Specifically, for a given target transcript $T_j$ and a retrieved candidate $T_k$, we define their category rank as $C(T_j) = \{c_{j,1}, c_{j,2},\cdots,c_{j,6}\}$ and 
$C(T_k) = \{c_{k,1}, c_{k,2},\cdots,c_{k,6}\}$, where each element represents the frequency of a parsing category in the respective transcript. 

The parsing similarity between $T_j$ and $T_k$ is then computed as:
\begin{equation}
\small
\begin{split}
\text{Sim}(T_j, T_k) &= \lambda_1 \cdot (1 - \textit{ND}(C(T_j), C(T_k))) \\
&+ \lambda_2 \cdot \textit{Cos}(R(T_j), R^{j}(T_k)),    
\end{split}
\end{equation}
where $\textit{Cos}(\cdot,\cdot)$ is the cosine similarity, $\textit{ND}(\cdot,\cdot)$ is the normalized position distance and $\lambda_1=\lambda_2=0.5$ are weighting coefficients. The order of $R^{j}(T_k)$ is aligned with rank $C(T_{j})$:
\begin{equation}
\small
\text{ND}(T_j, T_k) = \sum_{i=1}^6 \frac{|(\tau(C(T_j),i) - \tau(C(T_k),i)|}{i},
\end{equation}
where the index function $\tau:C \rightarrow \mathbb{R}^+$ assigns an index value to each element $c \in C$, $\tau(C(T_j),i) = 7-i$ and $\tau(C(T_k),i) = 7-t$, where $t$ \textit{ s.t. } $c_{k,i} = c_{j,t}$.

\paragraph{Ensemble In-Context Learners}  
To further enhance reasoning stability and mitigate label conflict, we introduce an ensemble learning strategy. Each learner processes a single demo retrieved via parsing similarity, and the final prediction is obtained via 3 learners majority voting.

\begin{table*}[!ht]
\centering
\small
\renewcommand{\arraystretch}{0.6} % 调整行间距
\setlength{\tabcolsep}{6pt} % 调整列间距
\caption{
\textbf{Main results on AD detection across three datasets.} Performance comparison of various fine-tuning, ICL, and our EK-ICL method on the Test, Lu, and Pitt. Metrics include Accuracy (Acc), Precision (Pre), Recall (Rec), and F1-score.
}
\begin{tabular}{l|c|cccc|cccc|cccc}
\toprule
\multirow{2}{*}{\textbf{Method}} & \multirow{2}{*}{\textbf{Shot}} & \multicolumn{4}{c|}{\textbf{Test} (\%)} & \multicolumn{4}{c|}{\textbf{Lu} (\%)} & \multicolumn{4}{c}{\textbf{Pitt} (\%)} \\
% \cmidrule(lr){3-6} \cmidrule(lr){7-10} \cmidrule(lr){11-14}
& & \textbf{Acc} & \textbf{Pre} & \textbf{Rec} & \textbf{F1} & \textbf{Acc} & \textbf{Pre} & \textbf{Rec} & \textbf{F1} & \textbf{Acc} & \textbf{Pre} & \textbf{Rec} & \textbf{F1} \\
\midrule[0.5pt]
\textbf{BERT$_c$} & N/A & 81.25 & 94.11 & 66.67 & 78.05 & 83.33 & 85.71 & 88.89 & 87.27 & 70.12 & 82.87 & 58.49 & 68.58 \\ %\cite{balagopalan2021bert}
\textbf{BERT$_f$} & N/A & 81.25 & 82.61 & 79.17 & 80.85 & 80.95 & 80.64 & 92.59 & 86.21 & 77.05 & 85.16 & 71.24 & 77.58 \\ %\cite{balagopalan2021bert}
\textbf{SPZ} & N/A & 91.66 & 95.45 & 87.50 & 91.30 & 83.33 & 83.33 & 92.59 & 87.72 & 77.96 & 88.07 & 69.93 & 77.96 \\ %\cite{spz}
\midrule
\textbf{LoRA$_{cls}$}  & N/A & 60.40 & 55.81 & \textbf{100} & 71.64 & 64.29 & 65.00 & \textbf{96.29} & 77.61 & 57.92 & 57.19 & \textbf{97.39} & 72.07 \\ %\cite{hu2022lora}
\textbf{LoRA$_{gen}$} & N/A & 41.67 & 43.33 & 54.17 & 48.15 & 47.62 & 60.87 & 51.85 & 56.00 & 51.18 & 57.19 & 49.35 & 52.98 \\ %\cite{hu2022lora}
\midrule
\multirow{4}{*}{\textbf{ICL$_{Van}$}} 
& 0 & 52.08 & 55.55 & 20.83 & 30.30 & 38.09 & 54.54 & 22.22 & 31.58 & 47.36 & 56.30 & 24.84 & 34.47 \\
& 1 & 64.58 & 73.33 & 45.83 & 56.41 & 45.23 & 70.00 & 25.93 & 37.84 & 50.27 & 61.38 & 29.08 & 39.47 \\
& 2 & 58.33 & 61.11 & 45.83 & 52.38 & 42.86 & 58.82 & 37.04 & 45.45 & 48.63 & 56.90 & 32.35 & 41.25 \\
& 3 & 52.08 & 52.63 & 41.66 & 46.51 & 59.52 & 75.00 & 55.56 & 63.83 & 55.74 & 65.52 & 43.46 & 52.26 \\
\midrule
\multirow{3}{*}{\textbf{ICL$_{Sem}$}} 
& 1 & 50.00 & 50.00 & 37.50 & 42.86 & 42.86 & 63.64 & 25.30 & 36.84 & 45.36 & 51.63 & 31.05 & 38.78 \\
& 2 & 45.83 & 44.44 & 33.33 & 38.09 & 38.09 & 53.33 & 29.63 & 38.09 & 46.08 & 51.96 & 43.14 & 47.14 \\
& 3 & 37.50 & 39.29 & 45.83 & 42.31 & 42.85 & 57.14 & 44.44 & 50.00 & 49.73 & 55.55 & 49.02 & 52.08 \\
\midrule
\multirow{4}{*}{\textbf{ICL$_{Log}$}  } %\cite{slm_logits}
& 0 & 52.08 & \textbf{100.00} & 4.17 & 8.00 & 40.48 & 75.00 & 11.11 & 19.35 & 49.73 & \textbf{100.00} & 9.80 & 17.86 \\
& 1 & 52.08 & \textbf{100.00} & 4.17 & 8.00 & 38.10 & \textbf{100.00} & 3.70 & 7.14 & 44.44 & 51.85 & 4.58 & 8.41 \\
& 2 & 52.08 & 60.00 & 12.05 & 20.69 & 40.48 & 66.67 & 14.81 & 24.24 & 47.72 & 67.27 & 12.09 & 20.50 \\
& 3 & 56.25 & 64.29 & 37.50 & 47.37 & 52.38 & 70.59 & 44.44 & 54.55 & 53.37 & 65.62 & 34.31 & 45.06 \\
\midrule
\multirow{4}{*}{\textbf{ICL$_{Ens}$} } %\cite{ensemble_icl}
& 0 & 45.83 & 40.00 & 16.67 & 23.53 & 45.24 & 70.00 & 25.93 & 37.84 & 45.54 & 52.29 & 26.14 & 34.86 \\
& 1 & 52.08 & 54.54 & 24.00 & 34.28 & 40.48 & 62.50 & 18.51 & 28.57 & 47.18 & 54.49 & 31.69 & 40.08 \\
& 2 & 43.75 & 42.28 & 37.50 & 40.00 & 33.33 & 47.37 & 33.33 & 39.13 & 43.53 & 49.15 & 37.90 & 42.80 \\
& 3 & 47.92 & 48.15 & 54.17 & 50.98 & 33.33 & 47.62 & 37.03 & 41.67 & 53.55 & 59.62 & 51.63 & 55.34 \\
\midrule
\textbf{Ours (EK-ICL)} & 1 & \textbf{93.75} & \textbf{100.00} & 87.50 & \textbf{93.33} & \textbf{88.09} & 92.31 & 88.89 & \textbf{90.57} & \textbf{80.51} & 94.61 & 68.95 & \textbf{79.77} \\
\bottomrule
\end{tabular}
\label{tab:full_results}
\end{table*}

\begin{table*}[!ht]
\centering
\small
\renewcommand{\arraystretch}{0.9}
\caption{Ablation results for the EK-ICL across Test dataset, Lu, and Pitt corpora. The symbol (-) indicates that Recall and F1 could not be computed due to a division-by-zero scenario.}
\begin{tabular}{l|cccc|cccc|cccc}
\toprule
\multirow{2}{*}{\textbf{Method}} &  \multicolumn{4}{c|}{\textbf{Test}} & \multicolumn{4}{c|}{\textbf{Lu}} & \multicolumn{4}{c}{\textbf{Pitt}} \\
& \textbf{Acc } & \textbf{Pre } & \textbf{Rec } & \textbf{F1 } & \textbf{Acc } & \textbf{Pre} & \textbf{Rec } & \textbf{F1 } & \textbf{Acc } & \textbf{Pre } & \textbf{Rec } & \textbf{F1 } \\
\midrule
\textbf{Ours}                  & 93.75 & 100   & 87.50  & 93.33 & 88.09 & 92.31 & 88.89 & 90.57 & 80.51 & 94.61 & 68.95 & 79.77 \\
\textbf{w/o Confidence Scores}            & 50.00 & 100   & -   & -  & 38.09 & 100   & 3.70  & 7.14  & 44.26 & 50.00 & 0.60  & 1.29  \\
\textbf{w/o Features Scores}          & 87.50 & 100   & 75.00  & 85.71 & 76.19 & 84.00 & 77.78 & 80.77 & 71.76 & 91.26 & 54.57 & 68.30 \\
\textbf{w/o Parsing Search}    & 83.33 & 100   & 66.67  & 80.00 & 83.33 & 91.67 & 81.48 & 86.27 & 71.22 & 91.57 & 53.26 & 67.35 \\
\bottomrule
\end{tabular}

\label{tab:ablation_result}
\end{table*}

\section{Experiment}
\subsection{Setup}
\subsubsection{Datasets}
We evaluated the proposed EK-ICL framework on three standard AD detection datasets: ADReSS challenge dataset \cite{luz2020alzheimer}, Lu \cite{lu} and Pitt corpora \cite{pitt}.

All datasets consist of transcripts formatted according to the CHAT protocol \cite{macwhinney2000childes}, ensuring consistency in pre-processing and facilitating cross-study comparisons.

\subsubsection{Baselines}
To comprehensively evaluate the performance of EK-ICL, we compared it with a range of baselines\footnote{SLM methods are based on bert-base-uncased, and LLM methods are based on llama3.1-8B-instruct. All experiments are conducted on one NVIDIA Quadro RTX 8000 48G GPU.}.%Detailed setup for LoRA is listed in Appendix~\ref{appendix:lora_details}. 

\paragraph{Fine-tuning Methods}
1) BERT (coarse-grained) \cite{balagopalan2021bert}: Utilizes the CLS token for classification;
2) BERT (fine-grained) \cite{balagopalan2021bert}: Extracts representations via GlobalMaxPooling;
3) SPZ \cite{spz}: Employs semantic perturbations and zonal-mixing to enhance robustness;
4) LoRA (CLS) \cite{hu2022lora}: Applies low-rank adaptation for parameter-efficient fine-tuning on LLM under classification task;
5) LoRA (GEN) \cite{hu2022lora}: Applies low-rank adaptation for parameter-efficient fine-tuning on LLM under text generation task.

\paragraph{ICL Methods}
6) Vanilla ICL: Basic ICL using randomly selected demonstrations;
7) Semantic ICL: Incorporates semantically similar demonstrations to improve prediction accuracy;
8) Logits ICL \cite{slm_logits}: Utilizing the logits of SLM to assist LLM;
9) Ensemble ICL \cite{ensemble_icl}: Combines multiple in-context learners and use the majority vote to obtain the final classification.

\begin{figure*}[!ht]
\centering
\small
\subfloat[]{%
  \includegraphics[height=5cm]{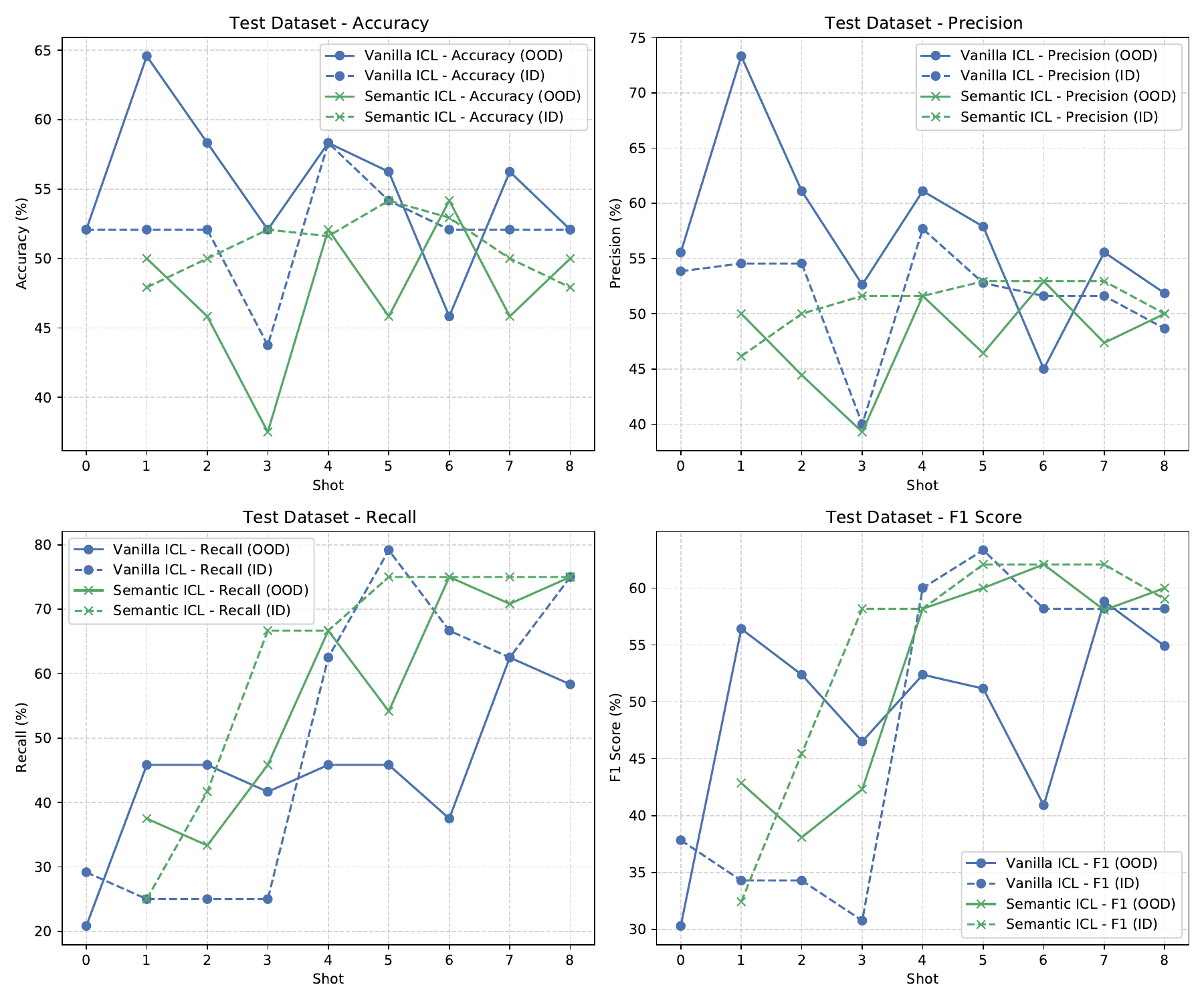}%
  \label{fig:ek_combo_a}}
\subfloat[]{%
  \includegraphics[height=5cm]{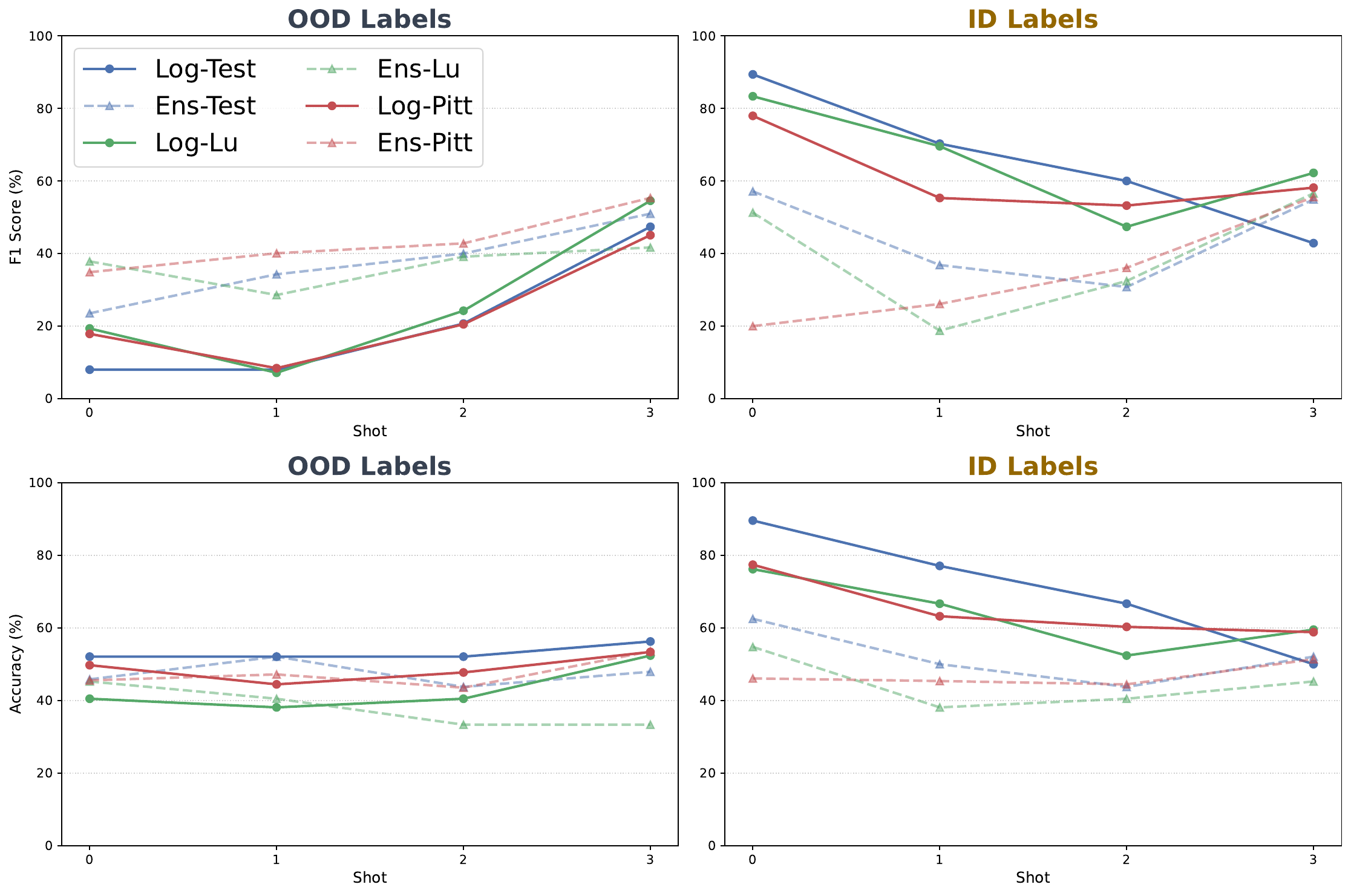}%
  \label{fig:ek_combo_b}}
\caption{\textbf{Comparative performance of ICL baselines under OOD/ID label-word settings.}
(a) ICL$_{\text{Van}}$ and ICL$_{\text{Sem}}$ on the Test dataset. 
(b) ICL$_{\text{Log}}$ and ICL$_{\text{Ens}}$ on three datasets.}
\label{fig:ek_combo}
\end{figure*}

\begin{figure*}[!ht]
\centering
\small
\includegraphics[width=0.85\linewidth]{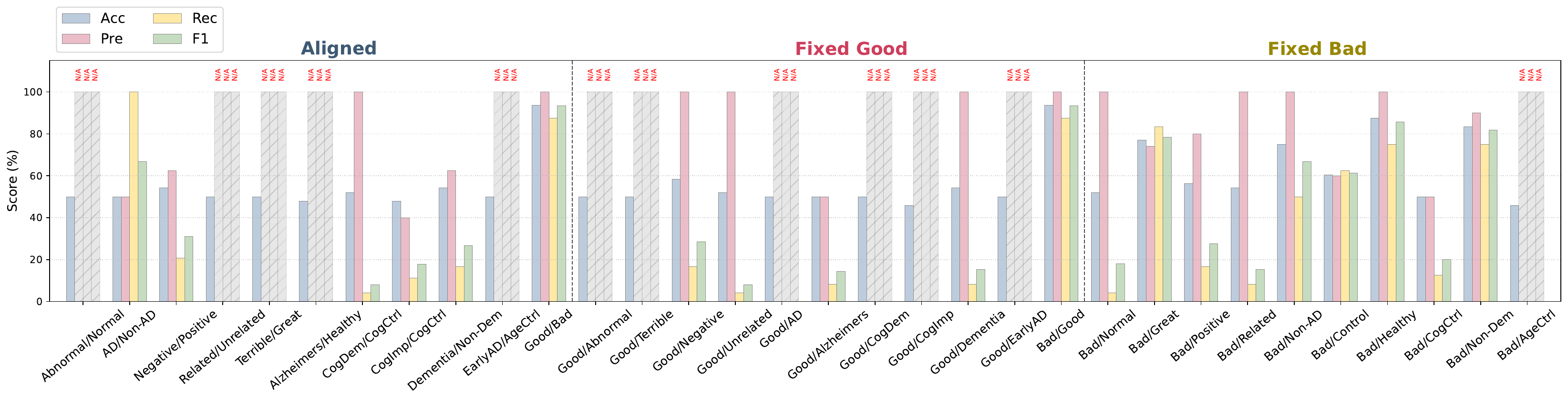}
\caption{\textbf{Performance of EK-ICL on the Test set across 30 label word pairs categorized into three configurations: \textbf{Aligned}, \textbf{Fixed Good}, and \textbf{Fixed Bad}}. Gray hatched bars indicate values that are not applicable due to division by zero.}\label{fig:label_pairs}
\vspace{-0.4cm}
\end{figure*}

\subsubsection{Evaluation Metrics}
We employed standard classification metrics (Accuracy, Precision, Recall, and F1-Score) to evaluate model performance. Results are reported across both balanced and imbalanced datasets to assess robustness and generalization.

\subsection{Main Results}
We evaluate EK-ICL for AD detection in both data-scarce and OOD scenarios, against FPFT, PEFT, and multiple ICL paradigms.

\textbf{FPFT on SLMs consistently achieve strong performance, but PEFT on LLMs  cannot.}
FPFT approaches such as SPZ yield top-tier results: 91.66\%/83.33\%/77.96\% accuracy on Test, Lu, and Pitt respectively. This demonstrates the capacity of SLMs, when fully fine-tuned, to capture fine-grained linguistic cues crucial for AD detection. By contrast, PEFT methods on LLMs (LoRA$_{cls}$) perform markedly worse (e.g., 60.40\% accuracy on Test), and text generation, oriented PEFT (LoRA$_{gen}$) further degrades. These results reveal the limitations of parameter-efficient tuning for subtle, low-resource clinical tasks.

\textbf{Standard ICL methods struggle in this domain.}
The best vanilla ICL ($shot=1$) achieves only 64.58\%/50.00\%/50.27\% accuracy on Test, Lu, and Pitt, with ICL$_{Sem}$ (semantic retrieval) performing even worse due to the semantic homogeneity of AD transcripts. Extensions leveraging logits (ICL$_{Log}$) or ensemble voting (ICL$_{Ens}$) bring no consistent gains, confirming that isolated optimization (logits, ensembles, or semantic retrieval alone) is insufficient for robust OOD adaptation.

\textbf{EK-ICL overcomes these limitations by integrating explicit knowledge into ICL.}
Our framework unifies SLM-derived confidence scores, parsing feature scores, and ID label replacement (e.g., Good/Bad), reinforced with a parsing-based demo retrieval and ensemble majority voting. This multi-faceted design ensures both task–label and context–task alignment. 93.75\%/88.09\%/80.51\% accuracy (93.33\%/90.57\%/79.77\% F1) on Test, Lu, and Pitt. These results highlight the necessity of simultaneously integrating explicit knowledge, structured retrieval, and label alignment for reliable LLM reasoning in AD detection. Our ablation confirms that none of these components alone, or in naive pairwise combination, can yield comparable gains.

\subsection{Ablation Study}
Our ablation study (shown Tab.~\ref{tab:ablation_result}), conducted under ID label conditions, reveals several key findings regarding the roles of SLM-logits based and parsing based EK modules in EK-ICL:

\textbf{SLM logits are indispensable.}
Removing the confidence scores leads to a dramatic drop in accuracy, highlighting their decisive and foundational role in calibrating LLM predictions. Nevertheless, as observed in the main results for ICL$_{Log}$, confidence scores alone do not yield stable improvements as demo counts increases.

\textbf{Parsing-based feature scores and search provide critical auxiliary benefits.}
While not as dominant as logits, parsing feature scores help the model capture fine-grained structural cues, and parsing-based demo search further improves retrieval quality by prioritizing syntactic diversity over semantic similarity. Removing either module results in clear performance degradation, confirming their additive and synergistic contributions.

\textbf{The effectiveness of EK is fundamentally constrained by label word selection.}
However, even with optimal EK integration, model performance is highly sensitive to the choice and alignment of label words. This is especially evident from the instability of ICL$_{Log}$ (see Tab.~\ref{tab:full_results}) in conventional label setting. These observations motivate our subsequent in-depth analysis: a systematic investigation of how different ID and OOD label word combinations, both aligned and non-aligned, affect EK-ICL performance.

\subsection{In-depth Analysis of Label Word Semantics}

To further understand how different ICL methods behave across diverse label words, we conducted extensive comparative analyses on the Test dataset (see Fig.~\ref{fig:ek_combo} and ~\ref{fig:label_pairs}). Our analyses yield three key insights:

\textbf{ID label words alone are insufficient without EK.}
As illustrated in Fig.~\ref{fig:ek_combo_a}, simply adopting ID-aligned labels (e.g., \textit{Good/Bad}) does not significantly enhance the performance of vanilla and semantic ICL methods in the absence of EK. These methods continue to struggle, indicating that \textit{semantic alignment with pre-trained LLM knowledge alone cannot adequately address the complex linguistic and structural challenges inherent to AD detection}.

\textbf{Logits and ensemble methods show limited effectiveness under OOD labels.}
ICL$_{Log}$ and ICL$_{Ens}$ methods under OOD label conditions exhibit consistently poor and stagnant performance, even with increasing demo counts (see Fig.~\ref{fig:ek_combo_b}). Under ID conditions, while the ensemble method remains ineffective, logits-based ICL initially achieves notable performance improvements at low demo counts. However, this advantage diminishes as more demos are introduced. This behavior underscores that \textit{confidence scores alone, without the additional structured parsing-based features and retrieval provided by EK-ICL, lack the stability needed for robust ICL performance in OOD and resource-scarce settings}.

\textbf{EK-ICL performance hinges on LLM-informed label pair alignment.}
As shown in Fig.~\ref{fig:label_pairs}, EK-ICL achieves its strongest performance with the \textit{Good/Bad} label pair, while other semantically aligned or clinically reasonable pairs (e.g., \textit{AD/Non-AD}) perform inconsistently or collapse. This suggests that effective label design is not merely about logical polarity or domain relevance, but about aligning with LLMs’ pre-trained semantic biases. In the \textbf{Aligned} and \textbf{Fixed Good} settings, performance is often dominated by the positive label, indicating LLMs’ preference for positive labels. In contrast, the \textbf{Fixed Bad} setting yields more stable and discriminative results across multiple pairings (e.g., \textit{Bad/Healthy}, \textit{Bad/Control}), implying that \textit{negative anchors like \textit{Bad} may provide stronger task-relevant separation in AD detection}. These results highlight that \textit{label effectiveness in OOD ICL settings depends on compatibility with LLM priors, which may deviate from human judgment}. A model-aware, empirically guided approach to label selection is therefore essential.

\section{Conclusion}
We propose EK-ICL, a novel in-context learning framework tailored for AD detection in data-scarce and OOD settings. EK-ICL explicitly injects three types of structured knowledge to overcome task recognition failure: (1) \textbf{confidence scores} derived from SLMs provide stable alignment between transcript and task objective; (2) \textbf{parsing feature scores} capture structural variations across transcripts, guiding informative demonstration selection; and (3) \textbf{label word replacement} aligns label semantics with LLM priors to mitigate label-task mismatch. Extensive experiments on three AD corpora show that EK-ICL outperforms both fine-tuning and standard ICL baselines. Further analysis reveals that robust AD detection requires joint alignment across label, context, and structure—dependencies that cannot be satisfied by any single knowledge source alone. Our findings underscore the importance of explicit knowledge in enabling LLMs to reason reliably in clinical OOD scenarios.

\section*{ACKNOWLEDGMENT}
This work was supported by the Key Research and Development Project of Hunan Province (No. 2025JK2119), Foundation of NUDT (HQKYZH2025KD004).

\bibliographystyle{IEEEtran}
\bibliography{custom}

\end{document}